\documentclass[11pt]{article}
\usepackage[margin=1in]{geometry}
\usepackage{amsmath}
\usepackage{amssymb}
\usepackage{graphicx}
\usepackage{booktabs}
\usepackage{authblk}
\usepackage{natbib}
\usepackage[colorlinks=true, citecolor=blue, linkcolor=blue, urlcolor=blue]{hyperref}

\title{Hard Negatives Reveal What Easy Negatives Hide: Cross-Lingual Harmfulness Representations Degrade with Resource Tier Under Hard Negatives}
\author[1]{Paras Balani}
\author[2]{Subhrakanta Panda}

\affil[1]{Department of Mathematics and Department of Computer Science, Birla Institute of Technology and Science, Pilani, Hyderabad Campus, Jawahar Nagar, Kapra Mandal, Medchal District, Telangana 500078, India}

\affil[2]{Department of Computer Science, Birla Institute of Technology and Science, Pilani, Hyderabad Campus, Jawahar Nagar, Kapra Mandal, Medchal District, Telangana 500078, India}
\date{}

\begin{document}
\maketitle

\begin{abstract}
Safety alignment in large language models is trained primarily in English, and recent work reports that the underlying harmfulness representation survives translation: English-trained probes separate harmful from harmless prompts almost as well in low-resource languages as in English. This has been taken as evidence that cross-lingual refusal failures mainly reflect calibration rather than representation quality. We show that this conclusion depends on the choice of negative examples. Across nine languages spanning three resource tiers, we replicate near-perfect transfer (AUROC $>0.98$) when harmless prompts come from an unrelated distribution (easy negatives). With XSTest contrast prompts, which are benign but surface-similar to harmful requests (hard negatives), transfer collapses in low-resource languages while remaining largely stable in high-resource languages. On Qwen2.5-7B-Instruct, mean AUROC drop increases from 0.003 in English to 0.017 in high-resource, 0.042 in mid-resource, and 0.276 in low-resource languages. The pattern replicates on Aya Expanse. Back-translation chrF controls and a matched-chrF comparison across three languages reduce the likelihood that translation quality explains the effect. The collapse remains after controlling for chrF (partial $r=0.70$, $p=0.03$). Tokenizer fertility correlates with the collapse and explains part of the resource-tier effect, but not all of it. The results show that easy-negative transfer can coexist with substantial degradation under hard negatives. Easy-negative evaluation alone therefore cannot establish that the harmfulness representation survives translation.
\end{abstract}

\section{Introduction and Related work}

Safety alignment in large language models is trained primarily in English, and it transfers imperfectly to other languages. \citet{yong2023lowresource} showed that translating harmful prompts into low-resource languages increases compliance in GPT-4. \citet{wang2505} extended this finding to activation space: a refusal direction extracted from English ablates refusal behavior in thirteen other languages with near-perfect effectiveness and is interchangeable across safety-aligned languages. More recently, \citet{oppong2608} tested four African languages directly in representation space and found that harmful prompts retain less than 10

A parallel line of work examines what safety probes actually read. \citet{zhao2507} showed that harmfulness and refusal are separate residual-stream directions: a model can judge a prompt harmless yet refuse, or judge it harmful yet comply. \citet{doda2605} found that probes trained on the final prompt token miss safety-relevant evidence that appears at earlier positions and later collapses before readout. These results indicate that probe performance depends strongly on where and how safety is read out.

Negative-example choice makes this dependency sharper. \citet{schwarz2607} constructed benign prompts matched to harmful prompts in topic and surface form and found that a mean-difference probe with AUROC 0.996-0.999 on disjoint sources falls to 0.590-0.690 on matched pairs across three model families. \citet{isaac2604} found the same pattern for a guard model: ShieldGemma-9B reaches AUROC 0.939 overall but only 0.304 TPR at 1\% FPR, with the largest gap on XSTest's hard-benign prompts. Neither paper tests this collapse across languages.

The closest antecedent is \citet{hanif2606}, who evaluated harmfulness-probe transfer across 23 languages and three model families. They found that a harmfulness direction extracted from high-resource activations separates harmful from harmless low-resource prompts almost as well as in English, even as behavioral refusal falls from 87.9\% to 43.9\%. They interpret this gap as calibration failure and show that resetting the decision threshold on a small number of target-language examples recovers much of the lost refusal rate.

We argue that this conclusion depends on the negative examples. \citet{hanif2606} use clearly harmless PolyRefuse prompts, the same type of easy negative that \citet{schwarz2607} and \citet{isaac2604} show can inflate AUROC through surface-form cues. Such negatives can make a representation appear intact while carrying little information about harmfulness once surface cues are controlled.

We test this by transferring an English-trained harmfulness probe to nine target languages using two negative sets: easy negatives, AdvBench against Alpaca instructions, matching prior cross-lingual work, and hard negatives, XSTest contrast prompts that are benign but resemble harmful requests. With easy negatives, transfer remains above 0.98 AUROC even for Amharic and Swahili, replicating the intact-representation result. With hard negatives, transfer collapses in the same languages, and the collapse increases as language-resource level decreases. High-resource languages lose less than 0.04 AUROC, whereas low-resource languages lose 0.20-0.42. A back-translation chrF control rules out translation quality as the sole explanation: Telugu and Vietnamese have matched chrF ($\approx$68) but differ five- to ten-fold in collapse magnitude, and the resource-tier effect survives partialling out chrF (partial $r = 0.70$, $p = 0.03$). The pattern replicates on Aya Expanse, with the same monotonic ordering and a comparable partial correlation.

Thus, the ``representation is present'' conclusion in \citet{hanif2606} depends on their use of easy negatives. Under hard negatives, the harmfulness representation itself degrades with language resource level, not only the decision boundary over it. We also test tokenizer fertility, defined as the number of subword tokens required per English word. Fertility correlates with collapse on both models ($r = 0.59$ and $r = 0.72$) and accounts for roughly half of the tier effect after inclusion in the partial correlation. With nine languages, however, neither factor clearly dominates. We therefore treat fertility as a correlate, not an established cause. \citet{datta2601} find that hallucination detectors degrade far less than task accuracy in low-resource languages, indicating that representational degradation is not automatic under low resource conditions and depends on the probed capability.

\section{Method}

\subsection{Data}

We construct two negative sets for a fixed pool of 200 harmful prompts and 250 harmless prompts, matched in size so that a probe trained on one set cannot exploit class imbalance.

Easy negatives pair AdvBench harmful behaviors \citep{zou2023advbench} with Alpaca instructions \citep{taori2023alpaca}, following the negative-set construction used in prior cross-lingual probing work. Hard negatives use XSTest \citep{rottger2024xstest}: the 200 harmful prompts are its contrast set, safe requests deliberately written to resemble unsafe ones, and the 250 harmless prompts are its remaining safe prompts. Both sets are sampled from the same underlying pools with a fixed seed, so the two conditions differ only in how surface form separates from meaning.

Before running either condition through a model we gate on a length check. A prompt-length-only classifier (negative token count as the decision score) scores 0.688 AUROC on easy negatives and 0.534 on hard negatives. Easy negatives clear this gate because AdvBench goals and Alpaca instructions differ systematically in length; XSTest's contrast pairs do not, since both halves are written to resemble each other. This is what makes XSTest suitable as the hard condition: a probe that separates its two halves cannot be doing so by reading length or imperative form.

We translate all 450 prompts from English into nine target languages spanning three resource tiers, using NLLB-200-distilled-600M \citep{nllb2022} with beam search (\texttt{num\_beams=4}, \texttt{max\_new\_tokens=256}). High-resource: Spanish, Mandarin, French. Mid-resource: Hindi, Vietnamese, Arabic. Low-resource: Telugu, Swahili, Amharic. Tier labels follow standard resource classifications for NLLB-200's training data; English is treated as the source language rather than assigned a tier.

To check whether translation quality alone could explain later results, we back-translate each language's hard-negative prompts to English through the same NLLB model and score chrF \citep{popovic2015chrf} against the English originals with sacrebleu. This gives a per-language chrF value that we later regress against the AUROC drop between conditions.

\subsection{Models}

We run the full pipeline on two open instruction-tuned models: Qwen2.5-7B-Instruct (28 transformer layers, 3584-dimensional residual stream) and Aya-Expanse-8B (32 layers, 4096-dimensional). Both load in 4-bit (NF4, double quantization, fp16 compute dtype) on a single T4 GPU. Neither model is fine-tuned or otherwise modified; all activations come from a single forward pass with no gradient computation.

\subsection{Probing setup}

For each prompt we apply the model's chat template with an empty assistant turn, tokenize with left padding, and run one forward pass with \texttt{output\_hidden\_states=True}. We take the last-token representation at every layer, including the embedding layer, giving one activation vector per prompt per layer. This is the last prompt token, not the last token of a generated response, so the probe reads what the model has computed about the input rather than what it is about to say. \citet{zhao2507} show this distinction matters: harmfulness and refusal are separate directions, with harmfulness encoded earlier and more robustly than the refusal decision that follows it. Our probe targets the former.

For each layer, we fit a logistic regression (\texttt{StandardScaler} plus \texttt{LogisticRegression}, $\ell_2$ penalty, $C=0.05$) on English activations only, using 5-fold cross-validation to score English itself and training on all English examples to score every other language. Regularization strength and the scaler matter here: with $n=450$ examples and $d \in \{3584, 4096\}$, the fit is deep in the underdetermined regime, and unscaled features make regularization behave erratically across layers. No target-language data is ever used to fit the probe; every non-English AUROC is a pure transfer number.

We select the peak layer separately for each condition, as the layer that maximizes cross-validated English AUROC. Easy negatives peak in early-to-mid layers; hard negatives peak substantially later, consistent with a real semantic concept requiring more computation than a surface cue. All reported non-English scores use the transfer probe evaluated at that condition's peak layer, with 95\% confidence intervals from a 5,000-resample stratified bootstrap over prompts. English confidence intervals use the cross-validated scores rather than an in-sample fit, since the latter is inflated to a point of being uninformative at $n=450$, $d>3000$.

\section{Results}

\subsection{Main result}

Table~\ref{tab:easy_vs_hard} reports transfer AUROC by language and negative-set condition, at the layer that maximizes cross-validated English AUROC in each condition (layer 13 for easy negatives, layer 23 for hard negatives, Qwen2.5-7B-Instruct). Under easy negatives, transfer holds above 0.98 AUROC for every language, including Amharic (0.977) and Swahili (0.991). This matches the pattern reported by \citet{hanif2606}: a harmfulness probe trained on English separates harmful from harmless prompts almost as well in low-resource languages as in English.

Under hard negatives, transfer collapses in the same low-resource languages. High-resource languages lose almost nothing: French drops from 1.000 to 0.993 (0.007), Spanish from 1.000 to 0.990 (0.010). Low-resource languages lose far more: Swahili drops from 0.991 to 0.786 (0.204), Telugu from 0.998 to 0.791 (0.208), Amharic from 0.977 to 0.561 (0.416). 

\begin{figure}[!ht]
\centering
\includegraphics[width=0.85\textwidth]{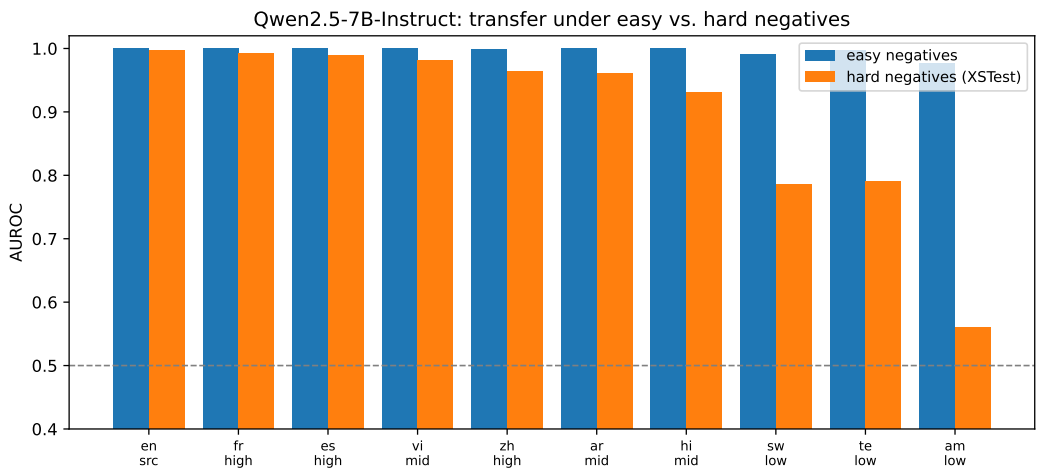}
\caption{Qwen2.5-7B-Instruct: transfer AUROC under easy negatives (AdvBench vs. Alpaca) and hard negatives (XSTest contrast prompts), by language and resource tier.}
\label{fig:easy_hard_qwen}
\end{figure}

\begin{table}[!ht]
\centering
\caption{Qwen2.5-7B-Instruct: transfer AUROC under easy negatives and hard negatives, by language and resource tier. Easy peak layer 13; hard peak layer 23.}
\label{tab:easy_vs_hard}
\begin{tabular}{lccccc}
\toprule
Language & Tier & chrF & Easy & Hard & Drop \\
\midrule
English    & source & 100.0 & 1.000 & 0.997 & 0.003 \\
French     & high   & 72.8  & 1.000 & 0.993 & 0.007 \\
Spanish    & high   & 76.6  & 1.000 & 0.990 & 0.010 \\
Vietnamese & mid    & 67.6  & 1.000 & 0.981 & 0.019 \\
Mandarin   & high   & 62.4  & 0.999 & 0.964 & 0.035 \\
Arabic     & mid    & 67.7  & 1.000 & 0.961 & 0.039 \\
Hindi      & mid    & 72.0  & 1.000 & 0.931 & 0.069 \\
Swahili    & low    & 61.9  & 0.991 & 0.786 & 0.204 \\
Telugu     & low    & 68.6  & 0.998 & 0.791 & 0.208 \\
Amharic    & low    & 55.5  & 0.977 & 0.561 & 0.416 \\
\bottomrule
\end{tabular}
\end{table}

The size of this drop tracks resource tier directly. Mean drop by tier: source (English) 0.003, high-resource 0.017, mid-resource 0.042, low-resource 0.276. Each tier's mean drop is roughly an order of magnitude larger than the tier above it.

Bootstrapped 95\% confidence intervals (5,000 resamples, stratified over prompts) separate low-resource languages from high-resource languages cleanly at the hard-negative layer: Swahili 0.786 [0.725, 0.843], Telugu 0.791 [0.666, 0.899], Amharic 0.561 [0.496, 0.601], against high-resource languages sitting at 0.96-0.99 with non-overlapping intervals.

The pattern replicates on a second model family. Table~\ref{tab:aya} reports the same easy-vs-hard comparison on Aya-Expanse-8B (32 layers, easy peak at layer 3, hard peak at layer 20). The tier ordering matches Qwen: mean drop by tier is 0.002 (source), 0.008 (high), 0.019 (mid), 0.251 (low). Per-language drops track closely between models: Amharic 0.416 (Qwen) vs. 0.365 (Aya), Swahili 0.204 vs. 0.216, Telugu 0.208 vs. 0.173.

\begin{figure}[!ht]
\centering
\includegraphics[width=0.85\textwidth]{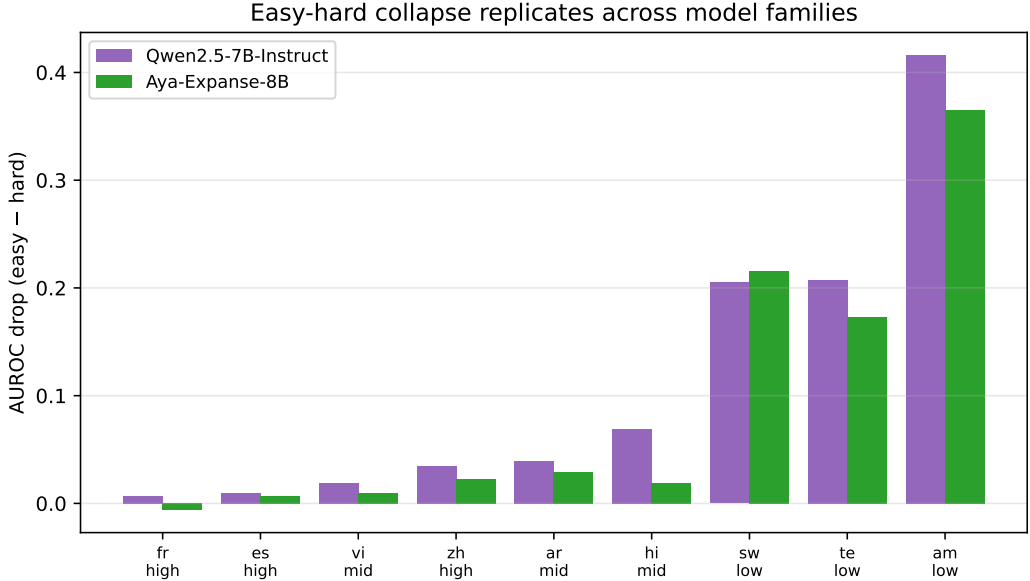}
\caption{Easy-to-hard AUROC drop by language, Qwen2.5-7B-Instruct and Aya-Expanse-8B. The tier ordering and approximate drop magnitude replicate across model families.}
\label{fig:replication}
\end{figure}

\begin{table}[!ht]
\centering
\caption{Aya-Expanse-8B: transfer AUROC under easy negatives and hard negatives, by language and resource tier. Easy peak layer 3; hard peak layer 20.}
\label{tab:aya}
\begin{tabular}{lccccc}
\toprule
Language & Tier & chrF & Easy & Hard & Drop \\
\midrule
English    & source & 100.0 & 1.000 & 0.998 & 0.002 \\
French     & high   & 72.8  & 0.989 & 0.995 & -0.005 \\
Spanish    & high   & 76.6  & 0.999 & 0.992 & 0.007 \\
Vietnamese & mid    & 67.6  & 1.000 & 0.990 & 0.010 \\
Hindi      & mid    & 72.0  & 0.999 & 0.980 & 0.019 \\
Mandarin   & high   & 62.4  & 0.997 & 0.974 & 0.023 \\
Arabic     & mid    & 67.7  & 0.994 & 0.965 & 0.029 \\
Telugu     & low    & 68.6  & 0.977 & 0.804 & 0.173 \\
Swahili    & low    & 61.9  & 0.983 & 0.767 & 0.216 \\
Amharic    & low    & 55.5  & 0.912 & 0.547 & 0.365 \\
\bottomrule
\end{tabular}
\end{table}

\subsection{Controls}

\paragraph{Translation quality.} A collapse concentrated in low-resource languages could reflect worse NLLB translation for those languages rather than worse model representation. We test this with back-translation chrF, computed once per language on the hard-negative prompt set and shared across both models since it depends only on the translations, not on which model is probed.

With Amharic included, chrF correlates strongly with the AUROC drop (Qwen: $r=-0.751$, $p=0.020$). Removing Amharic, the correlation collapses (Qwen: $r=-0.459$, $p=0.253$; not significant). Amharic has by far the worst chrF of the nine languages (55.5) and is the only point driving the raw correlation.

\begin{figure}[htbp]
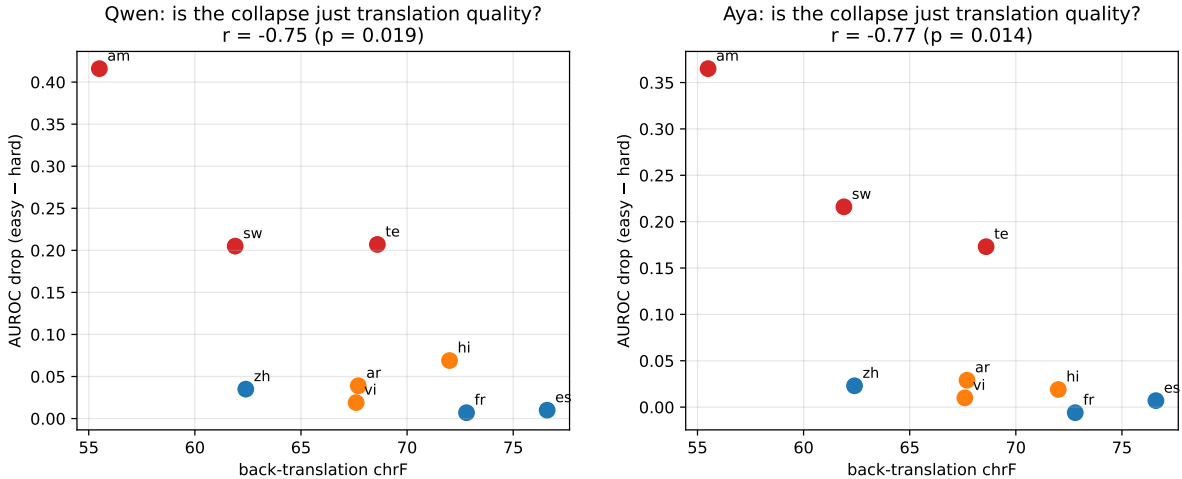

\centering
\includegraphics[width=0.48\textwidth]{fig_chrf_qwen.png}
\includegraphics[width=0.48\textwidth]{fig_chrf_aya.png}
\caption{AUROC drop against back-translation chrF, Qwen (left) and Aya (right). Amharic sits alone at low chrF and high drop; the remaining eight languages show no clear relationship between translation quality and collapse.}
\label{fig:chrf}
\end{figure}

The clearer test is a matched comparison at fixed chrF. Telugu, Arabic, and Vietnamese have almost identical chrF ($\approx$68) but very different drops: Telugu 0.208, Arabic 0.039, Vietnamese 0.019, a five- to ten-fold difference at matched translation quality. Since chrF is held constant across these three languages, translation quality cannot explain why Telugu collapses and Arabic and Vietnamese do not.

We confirm this with a partial correlation of resource tier against AUROC drop, controlling for chrF. On Qwen, partial $r(\text{tier}, \text{drop} \mid \text{chrf}) = 0.695$ ($p=0.038$); on Aya, partial $r = 0.709$ ($p=0.032$). Resource tier predicts the collapse after removing chrF's linear contribution, on both models.

\paragraph{Tokenizer fertility.} We test tokenizer fertility, the mean number of subword tokens per English word for the hard-negative prompts, as a candidate mechanism. Fertility is computed per model from that model's own tokenizer, using the same prompt set as the chrF control. Measured values: Qwen fertility ranges from 1.07 (Mandarin) to 8.19 (Telugu); Aya from 1.15 (Mandarin) to 9.19 (Telugu). Amharic is the second-highest for both (5.16 Qwen, 8.06 Aya).

\begin{figure}[!ht]
    \centering
    \includegraphics[width=0.9\textwidth]{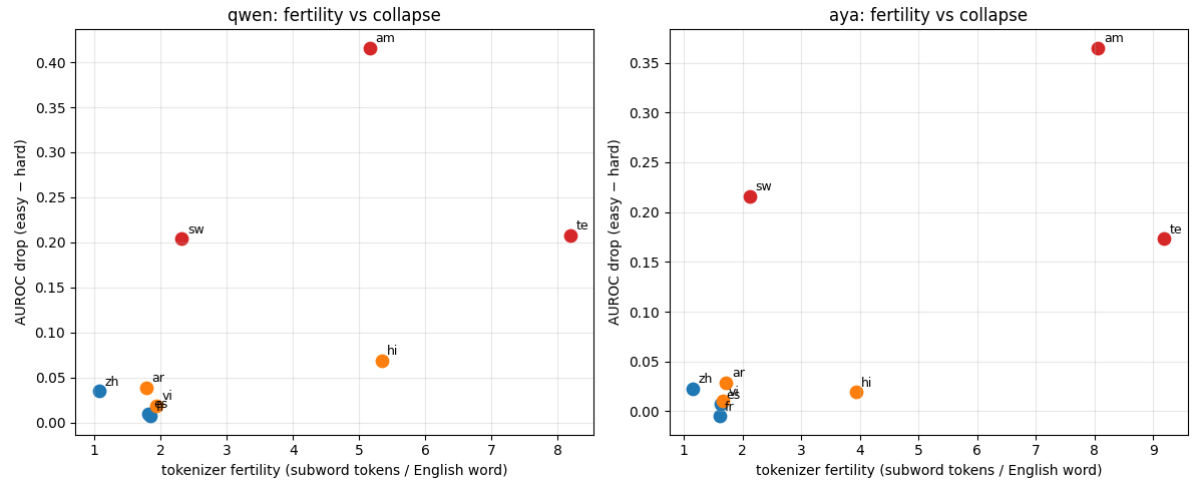}
    \caption{AUROC drop against tokenizer fertility. High-fertility languages cluster at high drop; Hindi is an outlier, with fertility comparable to Amharic but a far smaller drop.}
    \label{fig:fertility}
\end{figure}

Fertility correlates with the AUROC drop on both models: Qwen $r=0.592$ ($p=0.093$), Aya $r=0.720$ ($p=0.029$). Adding fertility to the partial correlation reduces the remaining tier effect: partial $r(\text{tier}, \text{drop} \mid \text{chrf}, \text{fert})$ falls to 0.280 ($p=0.465$) on Qwen and 0.419 ($p=0.261$) on Aya, both no longer significant. Fertility itself, controlling for chrF and tier, does not reach significance either: partial $r(\text{fert}, \text{drop} \mid \text{chrf}, \text{tier}) = 0.558$ ($p=0.118$) on Qwen, $0.547$ ($p=0.128$) on Aya. At nine languages, neither tier nor fertility cleanly dominates the other once both are included alongside chrF; fertility absorbs roughly half of the raw tier effect without eliminating it and without itself reaching significance as an independent predictor.

Hindi is the clearest exception to the fertility story: fertility of 5.34 (Qwen) or 3.93 (Aya), comparable to or above Amharic's 5.16/8.06, but a drop of only 0.069 (Qwen) or 0.019 (Aya), close to the mid-resource languages rather than the low-resource ones.

\section{Discussion}

 Near-perfect transfer with easy negatives, sharp degradation with hard negatives, a monotonic resource-tier gradient, and persistence after controlling for translation quality challenge the interpretation in \citet{hanif2606}. Their evaluation uses PolyRefuse harmless prompts that are clearly benign and differ in surface form from harmful prompts. Under this setup, harmfulness transfers well even to low-resource languages. \citet{hanif2606} interpret the result as evidence that the representation is intact and that the refusal gap is mainly a calibration problem. Our results show a different pattern with hard negatives. The same probing setup loses most of its separating power when surface-form shortcuts are removed, especially in low-resource languages.

The threshold reset in \citet{hanif2606} can therefore recover refusal behavior from a weak residual signal. A weak representation can still support useful thresholding. High easy-negative AUROC does not distinguish a weak representation from an intact one. \citet{schwarz2607} and \citet{isaac2604} show that easy-negative AUROC can remain high when probes rely on surface features instead of the underlying concept. Easy-negative transfer alone cannot establish that the representation is preserved.

 On Qwen, mean AUROC drop increases from 0.003 in English to 0.017 in high-resource, 0.042 in mid-resource, and 0.276 in low-resource languages. Aya shows a similar progression. The four-tier pattern is stronger than a simple high-resource versus low-resource comparison. The matched-chrF comparison among Telugu, Arabic, and Vietnamese also separates resource tier from translation quality. These three languages have similar chrF scores but show five- to ten-fold differences in collapse magnitude.

Fertility accounts for part of the variation but not all of it. Fertility correlates with collapse and explains roughly half of the raw tier effect. After controlling jointly for chrF and fertility, tier is no longer significant on either model. Fertility is also no longer significant after controlling for chrF and tier. Hindi provides a further constraint on a fertility-only explanation. Its fertility is comparable to or higher than Amharic's, but its collapse is much smaller and closer to the mid-resource languages. The data do not identify the remaining mechanism.

The results also matter for evaluating internal safety monitors. A probe tested only with easy negatives can appear effective across languages while losing most of its separating power on hard negatives in low-resource languages. XSTest-style contrast prompts are ordinary requests with wording that resembles unsafe requests. A monitor that fails on these negatives has not been adequately tested against cases that remove simple surface-form shortcuts.

\section{Solutions}

The hard-negative collapse points to two separate problems: evaluation and representation quality.

\paragraph{Evaluation.} A probe or monitor validated only with easy negatives may not predict performance in low-resource languages. Cross-lingual safety monitors should therefore be tested with negatives that match the harmful class in surface form. XSTest provides this construction in English. The same approach can be extended to other languages by translating or authoring matched benign prompts. This requires a change to the evaluation protocol, not the model.

\paragraph{The representation itself.} Evaluation can expose the degradation but cannot correct it. Two interventions follow from the results.

Tokenizer fertility provides one possible intervention. Low-resource languages with high fertility require more subword tokens per word, which may weaken the harmfulness signal at the final prompt token. Vocabulary augmentation or tokenizer retraining for languages such as Telugu, Amharic, and Swahili can test this hypothesis. The same probing pipeline can then measure whether lower fertility produces smaller AUROC drops.

Steering provides a second intervention. English-derived harmfulness directions may retain some signal in low-resource languages, as suggested by the above-chance bootstrap intervals for Swahili and Telugu. Adding these directions to the residual stream at inference time can test whether steering restores separation on hard negatives. A recovery would support a signal-strength explanation. No recovery, especially for Amharic where hard-negative AUROC is near chance, would provide evidence for a deeper representational deficit.

Both interventions can be tested with the existing pipeline. The current setup already provides layer-wise activations across languages and models, so neither intervention requires a new evaluation design.

\section{Conclusion}

Cross-lingual harmfulness probing has largely been evaluated with easy negatives, where harmful prompts are contrasted with harmless prompts from an unrelated distribution. Under this setup, English-trained probes transfer well to low-resource languages. This result supports the interpretation in \citet{hanif2606} that the harmfulness representation survives across resource tiers and that the refusal gap is mainly a calibration problem.

We replaced the harmless prompts with XSTest contrast prompts that match harmful prompts in surface form and reran the same probing pipeline on the same languages and models. Transfer remains strong in high-resource languages but collapses in low-resource languages. On Qwen2.5-7B-Instruct, mean AUROC drop increases from 0.003 in English to 0.017 in high-resource, 0.042 in mid-resource, and 0.276 in low-resource languages. Aya-Expanse-8B shows a comparable progression. Back-translation chrF controls and the matched-chrF comparison between Telugu, Arabic, and Vietnamese reduce the likelihood that translation quality explains the result. Tokenizer fertility correlates with the collapse and explains part of the tier effect, but Hindi shows that fertility alone cannot explain the pattern.

The harmfulness representation therefore appears intact under easy negatives but degrades substantially under hard negatives in low-resource languages. Easy-negative AUROC cannot distinguish an intact semantic representation from a surface-form shortcut. Cross-lingual safety representations should therefore be evaluated with hard negatives before claims of cross-lingual transfer are made.

\bibliographystyle{plainnat}
\bibliography{references}

\end{document}